\documentclass{IOS-Book-Article}

\usepackage{mathptmx}
\usepackage{soul}\setuldepth{article}
\def\hb{\hbox to 11.5 cm{}}

\usepackage{dsfont}
\usepackage{amsmath}
\usepackage{multirow}
\usepackage{array}
\usepackage{graphicx}

\usepackage{hyperref}
\usepackage{caption}

\begin{document}

\pagestyle{headings}
\def\thepage{}
\begin{frontmatter}              % The preamble begins here.

%\pretitle{Pretitle}
\title{Investigating Strategies for Clause Recommendation}

% \markboth{}{April 2022\hb}
%\subtitle{Subtitle}

\author[A]{\fnms{Sagar} \snm{Joshi}\thanks{Corresponding Author: Sagar Joshi; Email: sagar.joshi@resarch.iiit.ac.in}\orcid{0000-0001-6095-9713}},
\author[A]{\fnms{Sumanth} \snm{Balaji}\thanks{Corresponding Author: Sumanth Balaji; Email: sumanth.balaji@resarch.iiit.ac.in. Both the corresponding authors have contributed equally to this work.}\orcid{0000-0002-5669-7519}},
\author[A]{\fnms{Jerrin} \snm{Thomas}},
\author[B]{\fnms{Aparna} \snm{Garimella}} and \author[A]{\fnms{Vasudeva} \snm{Varma}}

\address[A]{\hyperref[sec:affiliation_a]{International Institute of Information technology, Hyderabad, India}}
\address[B]{Adobe Research, India}

\begin{abstract}
Clause recommendation is the problem of recommending a clause to a legal contract, given the context of the contract in question and the clause type to which the clause should belong.
With not much prior work being done toward the generation of legal contracts, this problem was proposed as a first step toward the bigger problem of contract generation.
As an open-ended text generation problem, the distinguishing characteristics of this problem lie in the nature of legal language as a sublanguage and the considerable similarity of textual content within the clauses of a specific type.
This similarity aspect in legal clauses drives us to investigate the importance of similar contracts' representation for recommending clauses.
In our work, we experiment with generating clauses for 15 commonly occurring clause types in contracts expanding upon the previous work on this problem and analyzing clause recommendations in varying settings using information derived from similar contracts.
\end{abstract}

\end{frontmatter}
% \markboth{April 2022\hb}{April 2022\hb}
%\thispagestyle{empty}
%\pagestyle{empty}

\section{Introduction}

\begin{figure}[h]
    \centering
    \includegraphics[scale=0.3]{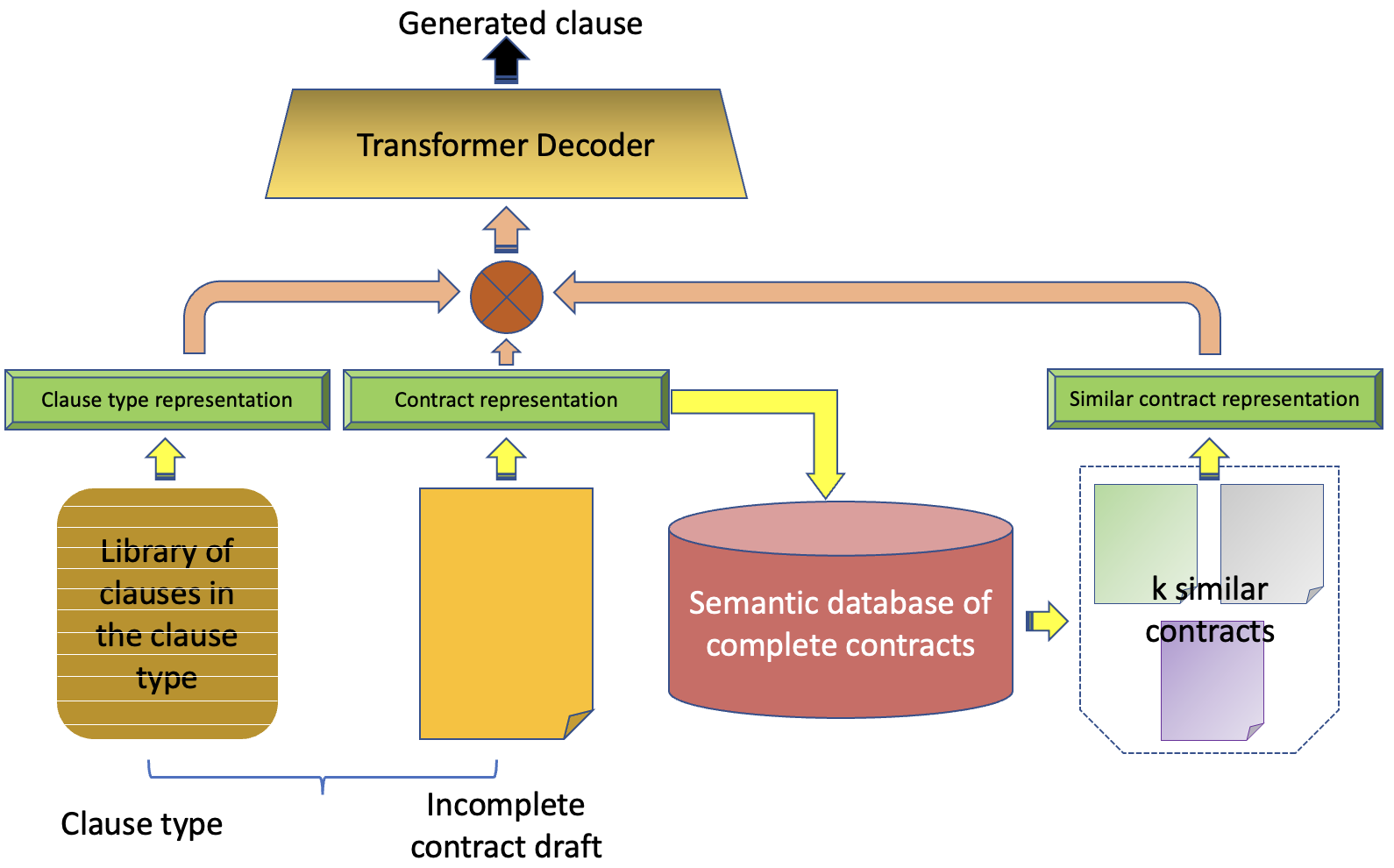}
    \caption{A pipeline for clause recommendation: We have an incomplete, in-draft contract at the input and the name of a clause type for which we need to recommend a new clause to the contract. We first compute the representation of the current contract. Since we know the clause type, we also compute a representation of the clause type based on a library of all the clauses under that type. Based on the representation of the current contract, we fetch $k$ similar contracts and average them to obtain a similar contract representation. Now, making use of these three representations, we aggregate them before sending them to our transformer-based clause decoder to generate or recommend a new clause of the specified clause type to be added to the in-draft contract.}
    \label{fig:pipeline}
\end{figure}

AI-driven assistance in drafting legal contracts as a tool can greatly benefit small and medium-sized enterprises. These enterprises often have limited legal support in contractual requirements compared to their large-scale counterparts that can afford sophisticated support from large legal teams. Contracts, being a type of legal document, can be characterized as being composed of a series of individual clauses or provisions, each capturing the legally binding rights, obligations, and agreements between the involved negotiating parties. As pointed out in [1], these clauses serve as the fundamental discourse units while drafting or reviewing contracts and encompass the legal essence of a contract.

The problem of clause recommendation [2] (ClauseRec) was proposed to assist contract drafting by recommending a clause to a contract. It was the first attempt at clause generation itself with the use of transformer-based techniques in NLP. For clause recommendation, consider an incomplete contract in the draft to which the drafter of the contract wishes to add a new clause of a specific clause type. Clause recommendation serves to add such a clause based by taking into account the context of the current contract in the process. The work also introduced the problem of clause type relevance prediction to determine the relevance of a clause type to be added to the contract before proceeding with clause generation. However, the problem was modeled as a binary classification problem to determine whether or not a given clause type is relevant to the current contract. The pipeline for clause recommendation nevertheless makes use of the clause type information, hence serving the problem of clause type relevance as a secondary problem - extra to clause recommendation. However, it keeps opening the question of whether or not the clause type information should be taken from the user (i.e., the contract drafter).

In our work, we study the problem of clause recommendation by experimenting with several strategies for representing the context for clause generation and showing the significance of clause type information for clause generation. We improve the pipeline proposed by [2] by adding similar contract representations to the context. Figure \ref{fig:pipeline} shows our best-performing pipeline for clause recommendation, which incorporates information from similar contracts in addition to the current contract and specified clause type. The generation is analyzed over 15 commonly different clause types in legal contracts. With research on legal clause generation currently in its infancy, we conclude by discussing the potential areas for improvement in clause generation. We open-source our code to ease future work on this problem\footnote{\url{https://github.com/sagarsj42/strategies_for_clause_recommendation}}.

% \begin{table}[t]
% \centering
% \scriptsize
% % \resizebox{0.5\textwidth}{!}{
% \begin{tabular}{p{7.2cm}}
% % \toprule
% In addition, the Credit Parties shall pay all Other Taxes to the relevant Governmental Authorities in accordance with applicable Law. The Credit Parties shall deliver to Administrative Agent official receipts or other evidence of such payment reasonably satisfactory to Administrative Agent in respect of any Other Taxes payable hereunder promptly after payment of such Other Taxes. \\
% % \bottomrule
% \end{tabular}
% \caption{An example \textit{Other Taxes} clause from a legal contract}
% \label{tab:sample_clause}
% \vspace{-0.15in}
% \end{table}

\section{Related Work}

Previous work in legal contracts has focused on understanding tasks in legal NLP directed towards easier review and analysis of contracts. Identification of entities in contracts is looked upon in [3], [4], the latter introducing a dataset of 179 contracts specific to lease agreements. The dataset introduced by [4] also provides labeled data for identifying potentially unfair clauses, a task attempted prior by [5]. [6] identify critical contract clauses using a set of Context-Free Grammar (CFG) rules. The task of semantically retrieving legal clauses from a library of contracts as a span identification problem was introduced in [7]. An expert annotation dataset for contract review is provided in [8]. LEDGAR - a dataset for large-scale multilabel classification was introduced in [1]. The sheer size of this dataset and the variety of clause types covered makes its applicability much beyond the task for which it was introduced, an example being the current work for clause generation.

Since legal language is different from the general open domain language on which most of the models are pretrained, special attention has been paid to developing models to give representations specialized to the domain. LegalBERT [9] introduces a family of BERT-based models for application on downstream legal tasks. [2] train a further pretrained version of BERT for use in context building for clause recommendation and show the superiority of the trained embeddings by showing distinguishable representations based on the clause type. ALeaseBERT trained in [4] is another task-specific pretrained model on legal contracts. We make use of LegalBERT representations in our work.

\section{Method}

The problem of clause recommendation can be modeled as a controlled text generation problem $P(y|context)$ in which we generate a clause $y$ given some contextual representation $context$. The $context$ representation can be modeled by making use of the inputs that can be made available from an incomplete version of the contract being drafted. Since a contract is essentially a collection of clauses, the representation of a contract can be obtained by averaging the representation of each $clause$ in the contract. The average representation serves as the contract-specific contextual input. Thus for a contract having $n_{clause}$ clauses, we can use a sequence encoder to get the representation of each clause before averaging to get the contract representation.
\begin{equation}
    contract = \frac{\Sigma_{clause} Encoder(clause)} {n_{clause}}
\end{equation}

The type of a legal clause is highly indicative of the clause since clauses under a single type are similar to each other. So, the clause type representation $clause \textunderscore type$ can be considered indicative in determining the characteristic text of the clause to be generated. For calculating this representation, we consider the library of all the clauses, $N_{clause\textunderscore type}$ occurring under the concerned clause type and compute an average of the individual clause embeddings.
\begin{equation}
    clause \textunderscore type = \frac{\Sigma_{clause} Encoder(clause)} {N_{clause\textunderscore type}}
\end{equation}

The $contract$ representation can be used to retrieve $k$ most similar contracts ($sim\_contracts$) from an index of all the contract representations. Each of the retrieved $k$ contracts can be used to provide additional context in two ways: 

(1) $full \textunderscore sim \textunderscore contr$: Using the representations calculated for an entire contract and averaging over all $k$ contracts
\begin{equation}
\begin{split}
    full\_sim\_contr = (\Sigma_{sim\_contract \in sim\_contracts} \\
    [ \Sigma_{clause \in sim\_contract} \frac{\Sigma_{clause} Encoder(clause)} {n_{clause}}]) / k
\end{split}
\end{equation}

(2) $clause \textunderscore sim \textunderscore contr$: Using only the representation of the clauses of clause type $t$ for which we need a clause recommendation. Here, we average the per-contract representations calculated for the clauses of the specified type before averaging over the $k$ contracts.

\begin{equation}
\begin{split}
    clause \textunderscore sim \textunderscore contr = (\Sigma_{sim\_contract \in sim\_contracts} \\ 
     \left[ \frac { \Sigma_{clause \in sim\_contract} \mathds{1} \{clause \: \epsilon \: t \} \cdot Encoder(clause) } { \Sigma_{clause \in sim\_contract} \mathds{1} \{clause \: \epsilon \: t \} } \right] ) / \: k
\end{split}
\end{equation}

By making use of these representations, we compute the $context$ vector in the following ways:
\begin{enumerate}
    \item ONLY\textunderscore CONTR: Using only the $contract$ representation tries to predict the clause in a clause type agnostic setting, i.e., the model has no information of which particular clause type it has to recommend a clause, making the task more difficult since the model has to predict the topic as well as the underlying content, and thus results in much poorer performance as compared to clause type aware outputs.
    \begin{equation}
        context = contract
    \end{equation}
    \item CONTR\textunderscore TYPE: This was the methodology adopted in ClauseRec and that takes into account both - $contract$ as well as $clause \textunderscore type$ representations.
    \begin{equation}
        context = (contract + clause \textunderscore type) / 2
    \end{equation}
    \item CONTR\textunderscore FULLSIM: In this experiment performed in a clause type agnostic setting, we complement the $contract$ representation with the $full \textunderscore sim \textunderscore contr$ to see the effect of similar contract representation when the clause type is not known.
    \begin{equation}
        context = \left[contract \; ; \; full \textunderscore sim \textunderscore contr\right]
    \end{equation}
    Here, $\left[. \: ; \: .\right]$ indicates concatenation of the vectors.
    \item CONTR\textunderscore TYPE\textunderscore FULLSIM: The original ClauseRec representation is augmented with full \textunderscore sim \textunderscore contr.
    \begin{equation}
    \begin{split}
        context = \left[(contract + clause \textunderscore type) / 2 \; ; \; full \textunderscore sim \textunderscore contr \right]
    \end{split}
    \end{equation}
    \item CONTR\textunderscore TYPE\textunderscore CLAUSESIM: Since we are aware of the clause type of clause to generate, $clause \textunderscore sim \textunderscore contr$ is used of to see if more specific information can help generate better clauses.
    \begin{equation}
    \begin{split}
        context = \left[ (contract + clause \textunderscore type) / 2 \; ; \; clause \textunderscore sim \textunderscore contr \right]
    \end{split}
    \end{equation}
\end{enumerate}

Once we have the strategy for computing the $context$ vector, a language model with trainable parameters $\theta$ is trained to condition on this representation by minimizing the negative log-likelihood loss between the predicted and the expected output tokens.
\begin{equation}
    l_{gen} = -log[ \, p(y \: | \: context, \theta) \, ]
\end{equation}

\section{Experimentation}

This section explains our filtering and re-purposing of the LEDGAR dataset, experiments with encoder models, different values of $k$ for similar contract retrieval, and metrics chosen for evaluation.

\textbf{Dataset.} We re-purpose the LEDGAR [1] dataset for multilabel clause type identification for our task. The dataset in its cleaned version consists of 60,540 contracts having 846,274 clauses from 12,608 different types. We filter the dataset first to eliminate all contracts with less than five clauses present, resulting in 34,442 contracts, following which we select the top 15 clause types for analyzing the generation. The selected clause types, along with their clause counts and length information, can be seen in Table \ref{tab:ctdistri}.

\begin{table*}[ht]
    \centering
    \begin{tabular}{c|c|c|c}
        \hline
        Clause type & \# clauses & mean length & std length \\
        \hline
        governing laws & 15291 & 103.00 & 104.80 \\
        amendments & 12571 & 127.19 & 119.26 \\
        entire agreements & 11023 & 98.01 & 64.50 \\
        counterparts & 10415 & 80.47 & 55.49 \\
        notices & 9726 & 148.20 & 105.62 \\
        waivers & 8945 & 133.93 & 107.43 \\
        severability & 8776 & 107.72 & 61.93 \\
        expenses & 8365 & 138.66 & 119.72 \\
        successors & 8184 & 116.60 & 87.98 \\
        survival & 6102 & 89.14 & 84.58 \\
        assigns & 6099 & 106.94 & 82.61 \\
        assignments & 5976 & 127.23 & 95.94 \\
        representations & 5373 & 136.16 & 102.57 \\
        warranties & 5320 & 138.06 & 138.13 \\
        taxes & 5184 & 164.87 & 126.78 \\
        \hline
    \end{tabular}
    \caption{Distribution of the selected clause types from LEDGAR.}
    \label{tab:ctdistri}
\end{table*}

\textbf{Encoders.} Considering the domain-specific nature of English in legal contracts, directly using transformer encoder models pretrained on generic corpora would not yield good results, as remarked in [2]. We experimented across three models - LegalBERT [9] trained on diverse legal texts (LegalBERT-all), LegalBERT trained only on US contracts from EDGAR (LegalBERT-contracts), and a further pretrained version of BERT [10] on our task-specific data using masked language modeling objective (BERT-mlm). The further pretrained BERT model was pretrained for two epochs on the clauses in LEDGAR data. Unlike the previous work, we did not pretrain a ContractBERT model due to computational limitations and focused on using existing large pretrained models available. \texttt{base} versions of all the BERT-based models were used in the experiments performed. To determine the best encoder model for representation, we trained using the representations from these three models on CONTR\_TYPE, CONTR\_TYPE\_FULLSIM and CONTR\_TYPE \textunderscore CLAUSESIM strategies, the results of which are shown in Table \ref{tab:encperf}.

\vspace{-10pt}
\begin{table*}[h]
    \centering
    \begin{tabular}{|c|c|c|c|c|c|c|}
        \hline
        \multirow{3}{*}{Encoder} & \multicolumn{6}{c|}{Strategy} \\
        \cline{2-7}
        & \multicolumn{2}{c|}{CONTR\textunderscore TYPE} & \multicolumn{2}{c|}{CONTR\textunderscore TYPE \textunderscore FULLSIM} & \multicolumn{2}{c|}{CONTR\textunderscore CLAUSESIM} \\
        \cline{2-7}
        & ROUGE-L & BLEU & ROUGE-L & BLEU & ROUGE-L & BLEU \\
        \hline
        BERT-mlm & 38.39 & 22.46 & 38.50 & 22.33 & 38.33 & 21.49 \\
        LegalBERT-contracts & 38.31 & 21.53 & 38.89 & 22.01 & 37.45 & 20.63 \\
        LegalBERT-all & \textbf{38.75} & \textbf{22.96} & \textbf{39.28} & \textbf{22.69} & \textbf{38.85} & \textbf{23.33} \\
        \hline
    \end{tabular}
    \caption{Performance of encoder models on 3 strategies: CONTR\textunderscore TYPE, CONTR\textunderscore TYPE\textunderscore FULLSIM, CONTR\textunderscore CLAUSESIM.}
    \label{tab:encperf}
\end{table*}

LegalBERT trained on diverse legal corpora was consistently observed to outperform the other two models and was chosen as the base encoder for all subsequent experiments.

\textbf{Indexing.} An HNSW [11] index of the precomputed contract representations was built using FAISS [12] indexing library, and a similarity search was performed using L2 distance. Care was taken to eliminate the contract for which the recommended clause is not in retrieved contracts, and the similar contract representation was averaged on the rest. Because of this consideration, the minimum value of $k$ was kept as 2.

To understand the number of retrieved similar contracts needed to build a suitable context, the experiments were conducted by varying values of $k$ from 2 to 12 in steps of 2 for CONTR\textunderscore TYPE\textunderscore FULLSIM strategy. Not much variance in the text generations scores was observed, with the BLEU and ROUGE-L metrics staying more or less around the same values. The value of $k$ was fixed to 6 for subsequent experiments.

\textbf{Model.} A transformer decoder [13] with three layers was trained from scratch across all the experiments, following the setting in ClauseRec. A wordpiece tokenizer was trained on output clauses with a vocabulary size of 8192 tokens. For the final experiments, all the decoder models were trained for 50 epochs using AdamW [14] optimization with a learning rate schedule having 25\% warmup up to 6e-5 followed by linear decay. A batch size of 24 with accumulated gradients over three steps was used. Each of the experiments was performed using 2 RTX 2080 Ti GPUs.

Computing the input representations to the decoder was computationally heavy at training time, even without flowing gradients through the encoders for backpropagation. One epoch took approximately 17 hours on a single GPU, most of which was spent calculating the input representation. This prohibitively high time severely restricted the number of experiments that could be performed on our computational resources. Hence, to fasten the decoder training process, we serialized all the representations to be used prior to the experiment, which reduced the per-epoch per-GPU duration to less than 30 min.

\textbf{Metrics.} Commonly used text generation metrics ROUGE [15] and BLEU [16] were used to evaluate the system's performance. ROUGE, a recall-oriented metric, can be used to indicate how many of the necessary legal phrases are generated in clauses while BLEU is indicative of the precision of the phrases. Human evaluation was not performed due to practical constraints in obtaining domain-specific experts for clause comprehension.

\section{Results}

\vspace{-10pt}
\begin{table*}[h]
    \centering
    \begin{tabular}{|c|c|c|c|c|c|c|c|}
        \hline
        \multirow{2}{6em}{Strategy} & \multirow{2}{5em}{Clause type / Overall} & \multicolumn{3}{c|}{ROUGE} & \multicolumn{3}{c|}{BLEU} \\
        \cline{3-8}
        & & ROUGE-1 & ROUGE-2 & ROUGE-L & BLEU-1 & BLEU-2 & BLEU \\ 
        \hline
        \multirow{4}{6em}{ONLY\textunderscore CONTR} & governing laws & 38.44 & 18.89 & 30.55 & 14.84 & 11.46 & 10.22 \\
        & amendments & 33.25 & 13.78 & 24.35 & 15.78 & 12.45 & 12.18 \\
        & entire agreements & 34.26 & 13.68 & 25.13 & 10.88 & 10.12 & 9.60 \\
        & Overall & 34.94 & 15.08 & 25.52 & 13.49 & 12.23 & 11.32 \\
        \hline
        \multirow{4}{6em}{CONTR\textunderscore TYPE} & governing laws & 52.92 & 37.78 & 46.38 & 47.90 & 33.04 & 31.32 \\
        & amendments & 37.98 & 20.38 & 31.72 & 14.16 & 8.43 & 8.01 \\
        & entire agreements & 48.48 & 28.49 & 40.78 & 26.28 & 15.02 & 13.53 \\
        & Overall & 47.93 & 29.20 & 38.75 & 39.01 & 24.26 & 22.96 \\
        \hline
        \multirow{4}{6em}{CONTR \textunderscore FULLSIM} & governing laws & 40.57 & 22.11 & 33.48 & 27.45 & 15.24 & 14.70 \\
        & amendments & 28.62 & 7.85 & 19.23 & 20.13 & 6.58 & 5.61 \\
        & entire agreements & 23.78 & 5.25 & 16.12 & 10.97 & 2.41 & 1.25 \\
        & Overall & 31.45 & 12.43 & 23.30 & 21.06 & 8.60 & 8.14 \\
        \hline
        \multirow{4}{6em}{CONTR\textunderscore TYPE \textunderscore FULLSIM} & governing laws & 54.27 & 38.07 & 47.00 & 50.60 & 34.27 & 32.37 \\
        & amendments & 41.35 & 23.86 & 34.74 & 20.72 & 13.19 & 12.78 \\
        & entire agreements & 49.21 & 29.68 & 41.57 & 24.85 & 15.14 & 13.94 \\
        & Overall & \textbf{48.38} & \textbf{29.72} & \textbf{39.26} & \textbf{39.04} & \textbf{24.39} & \textbf{23.05} \\
        \hline
        \multirow{4}{6em}{CONTR\textunderscore TYPE \textunderscore CLAUSESIM} & governing laws & 52.66 & 36.62 & 45.89 & 47.84 & 33.24 & 31.66 \\
        & amendments & 40.54 & 21.42 & 32.12 & 24.37 & 14.63 & 14.16 \\
        & entire agreements & 49.79 & 30.17 & 42.00 & 28.21 & 17.58 & 16.26 \\
        & Overall & 47.75 & 29.30 & 38.85 & 39.12 & 24.60 & 23.33 \\
        \hline
    \end{tabular}
    \caption{Clause recommendation results across all the strategies tried out shown for top 3 clause types (based on no. of clauses) and overall metrics calculated on all the types.}
    \label{tab:res}
\end{table*}

Table \ref{tab:res} shows the metrics across all the strategies for the top 3 clause types (based on the no. of clauses in the dataset) and the overall scores computed for clauses for all the 15 clause types.

From the results, it can be easily observed that similar contracts with clause type and input contract provide the best contextual representation among the strategies tried. However, the strategy excluding similar contract representation performs competitively with the former. In a clause-type agnostic setting, the problem becomes much more difficult with modest ROUGE scores and dismal BLEU scores indicating a high amount of hallucinated, irrelevant content. Unlike clause type aware setting, the augmentation of similar contract information does not help, with an observed degradation in model output.

Among the top 3 clause types shown, the highest values are achieved by \textit{governing laws} followed by \textit{entire agreements} and \textit{amendment} for most cases. The relatively less value of BLEU scores indicates that models struggle to produce content with a high precision, and thus may have a substantial amount of hallucinated or irrelevant content.

\section{Analysis}

We conduct analysis based on the generation of clauses from the best performing CONTR\_TYPE\_FULLSIM strategy.

\begin{figure}[h]
    \centering
    \includegraphics[scale=0.4]{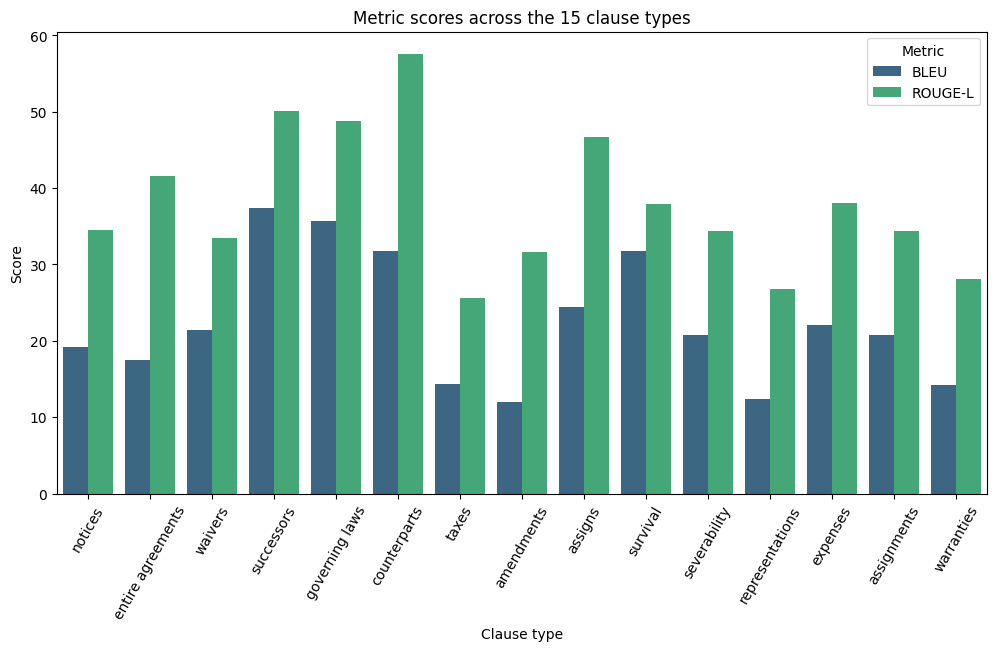}
    \caption{Variation of BLEU and ROUGE score across all 15 clause types considered using the CONTR\_TYPE\_FULLSIM}
    \label{fig:bleu-and-rouge-15types}
\end{figure}

The performance of clause generation across all the 15 clause types on BLEU and ROUGE-L scores can be seen in Figure \ref{fig:bleu-and-rouge-15types}. The clause types \textit{successors} and \textit{counterparts} have the highest BLEU and ROUGE-L scores, respectively. It can be observed that the clause types with generally higher BLEU (\textit{governing laws, counterparts, successors}) scores follow suit in ROUGE-L scores, and those performing poorly in terms of BLEU (\textit{amendments, representations, warranties, taxes}) have poor ROUGE-L scores.

\begin{figure}[h]
    \centering
    \includegraphics[scale=0.3]{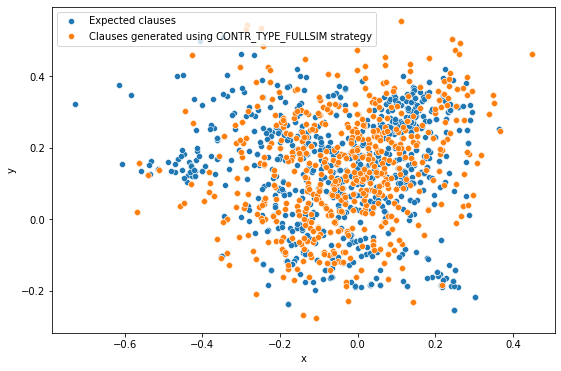}
    \caption{TSNE plot comparing the representations obtained for clause generated by the CONTR\_TYPE\_FULLSIM strategy against actual clauses}
    \label{fig:tsne-act-best}
\end{figure}

In Figure \ref{fig:tsne-act-best}, we plot the TSNE representations of the actual and generated clauses for a subset of 1000 randomly sampled clauses from our test set. We use LegalBERT-all for encoding the clauses, followed by dimensionality reduction to obtain the plot. The plot showcases the semantic closeness of the actual and generated clauses based on the closeness of the two distributions.

\begin{figure}[h]
    \centering
    \includegraphics[scale=0.25]{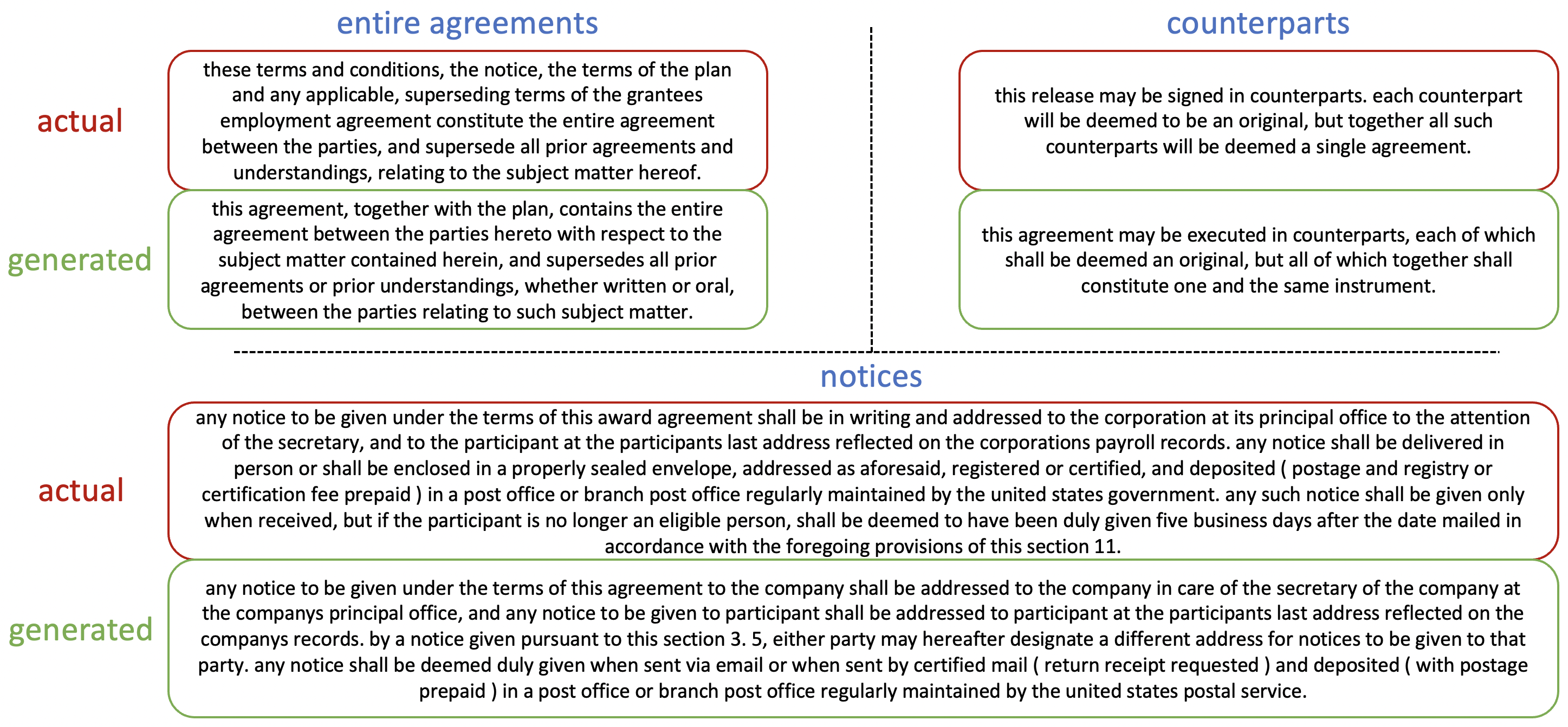}
    \caption{Examples of generated clauses in comparison to their expected (actual) counterparts using the CONTR\_TYPE\_FULLSIM strategy}
    \label{fig:outputs}
\end{figure}

We show examples of a few generated clauses by the model in Figure \ref{fig:outputs}. Based on the comparison with their expected counterparts, we can appreciate the ability of the model to generate characteristic clause content based on the type of clause. The exactness of the clause is not guaranteed here, as the model may generate semantically equivalent content deemed to be an equally valid clause, such as ``this release" v/s ``this agreement." Tailoring clauses to bear exact phrasal content can entail incorporating keyword-level information as future work.

One issue we would like to point out in the current generation is the tendency of the model to typically generate verbose clauses, which might have resulted from optimization on the training objective to include as many valid phrases as possible in a clause. As seen in Table \ref{tab:stats_lens}, the mean and median lengths of generated clauses are far more than the actual clauses. The Pearson correlation coefficient between the two length distributions turned out to be 0.36, indicating a weak correlation. Reducing such extraneous content in generated clauses will work towards finer clause generation.

\begin{table}[h]
    \centering
    \begin{tabular}{c|c|c|c}
         & mean & std & median \\
        \hline
        actual & 88.05 & 88.54 & 54.0 \\
        generated & 108.02 & 84.14 & 84.0 \\
    \end{tabular}
    \caption{Statistics of generated and actual clause lengths}
    \label{tab:stats_lens}
\end{table}

\section{Conclusion \& Future scope}
In this work, we explored the problem of clause recommendation by experimenting with several strategies for modeling the contextual input for recommendation and observed the effectiveness of similar contract representation in a clause-type aware setting. While decent results are obtained in this paradigm, the clause type agnostic setting remains a complex problem to solve and is more relevant for scaling clause recommendation to a broader set of clauses. With the considerably large size of contractual documents, future work can explore tackling the cold start problem in clause recommendation (i.e., recommending clause without prior contract state as context) and allowing for disentangling and customization of named entities in recommended clauses. Future work can also achieve robustness by handling a more diverse range and a larger number of clause types compared to the 15 clause topics we focus on.

\end{document}